\def\year{2022}\relax
\documentclass[letterpaper]{article} 
\usepackage{aaai22}  
\usepackage{times}  
\usepackage{helvet}  
\usepackage{courier}  
\usepackage[hyphens]{url}  
\usepackage{graphicx} 
\usepackage{subcaption}
\usepackage{amsmath}
\graphicspath{ {./figures/} }
\usepackage{natbib}  
\usepackage{caption} 
\DeclareCaptionStyle{ruled}{labelfont=normalfont,labelsep=colon,strut=off} 
\usepackage{algorithm}
\usepackage{algorithmic}

\title{LatentGAN Autoencoder: Learning Disentangled Latent Distribution}
\author{
    Sanket Kalwar,
    Animikh Aich, 
    Tanay Dixit,
    Adit Chhabra
}
\affiliations{
     Wobot Intelligence Inc.\\
     1119, Forrestal Lane \\
	 Foster City, CA 94404 \\

    sanket.kalwar@wobot.ai, animikh@wobot.ai, tanay@wobot.ai, adit.chhabra@wobot.ai
}

\usepackage{bibentry}

\begin{document}
\maketitle
\begin{abstract}
In autoencoder, the encoder generally approximates the latent distribution over the dataset, and the decoder generates samples using this learned latent distribution. There is very little control over the latent vector as using the random latent vector for generation will lead to trivial outputs. This work tries to address this issue by using the LatentGAN generator to directly learn to approximate the latent distribution of the autoencoder and show meaningful results on MNIST, 3D Chair, and CelebA datasets, an additional information-theoretic constrain is used which successfully learns to control autoencoder latent distribution. With this, our model also achieves an error rate of 2.38 on MNIST unsupervised image classification, which is better as compared to InfoGAN and AAE.
\end{abstract}
\section{Introduction}
Generative models like GAN\cite{goodfellow2014generative} and VAE\cite{kingma2014autoencoding} have shown remarkable progress in recent years.Generative adversarial networks have shown state-of-the-art performance in a variety of tasks like Image-To-Image translation\cite{isola2018imagetoimage}, video prediction\cite{liang2017dual}, Text-to-Image translation\cite{zhang2017stackgan}, drug discovery\cite{hong2019molecular}, and privacy-preserving\cite{shi2018ssgan}. VAE has shown state-of-the-art performance in a variety of tasks like image generation\cite{gregor2015draw}, semi-supervised learning\cite{maale2016auxiliary}, and interpolating between sentences\cite{bowman2016generating}. VAE approximates its latent distribution by ELBO method and uses Gaussian or uniform distribution as a marginal latent distribution which leads to huge reconstruction error. Adversarial autoencoder(AAE)\cite{makhzani2016adversarial} uses adversarial training to train VAE, and LatentGAN is directly inspired by the AAE strategy of training VAE. $\beta$-VAE\cite{burgess2018understanding} can be used for learning disentangled representation, but it needs $\beta$ hyperparameter search and also huge $\beta$ would lead to huge reconstruction error. InfoGAN\cite{chen2016infogan} have shown to learn disentangled representation in GAN. It learns to disentangle representation by maximizing the mutual information between a subset of generated sample and the output of the recognition network. Learning disentangled representation and understanding generative factors might help in a variety of tasks and domains\cite{6472238,burgess2018understanding}. Disentangled representation can be defined as one where single latent units are sensitive to changes in single generative factors, while being relatively invariant to changes in other factors. Disentangled representations could boost the performance of state-of-the-art AI approaches in situations where they still struggle but where humans excel\cite{lake2016building}.
In this paper, we have addressed the issue of learning disentangled control by learning the latent distribution of autoencoder directly using LatentGAN generator, and LatentGAN discriminator which tries to discriminate whether the sample belongs to real or fake latent distribution.  Additionally, we also use mutual information inspired directly from InfoGAN to learn disentangled representation.
\section{Related Work}
In this work, we present a new way to learn control over autoencoder latent distribution with the help of AAE  \cite{makhzani2016adversarial} which approximates posterior of the latent distribution of autoencoder using any arbitrary prior distribution and using \cite{chen2016infogan} for learning disentangled representation. The previous work by \cite{wang2019learning} had used a similar method of learning the latent prior using AAE along with a perceptual loss and Information maximization regularizer to train the decoder with the help of an extra discriminator.
\section{Contribution}
In this work, we are trying to learn control over autoencoder and in doing so following are our contribution:
\begin{itemize}
\item We have shown that it is possible to approximate autoencoder latent distribution directly using LatentGAN generator $G$ without using extra discriminator as in \cite{wang2019learning}.
\item We are also able to learn disentangled control over autoencoder latent distribution by using same LatentGAN Discriminator $D$ and have shown some results in the experiment section.	
\end{itemize}
And in doing so we are also able to get 2.38 error rate on MNIST unsupervised classification which is far less then InfoGAN and Adversarial autoencoder.
\section{Method}
\begin{figure}[hbt!]
	\includegraphics[scale=0.4]{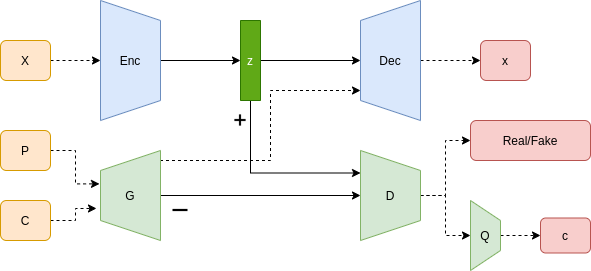}
	\caption{LatentGAN Autoencoder model}
	\label{fig:latentgan}
\end{figure}

We train autoencoder i.e. $Dec(Enc(X)) \sim f_{autoencoder}(X)$ and LatentGAN generator $G$ and discriminator $D$ simultanously,this helps generator to compete with discriminator.Autoencoder training objective is as follows:

\[
	L_{autoencoder} = \|X-f_{autoencoder}(X)\|^2   \tag{1} \label{eq:1}
\]
LatentGAN tries to model data distribution $x \sim p_{data}(z)$ where $z$ is the latent embedding of the autoencoder, and $lc$ denotes latent code which generated by concatenating Gaussian $P$ and code $c$, as show in the Figure \ref{fig:latentgan}. LatentGAN training objective is as follows:

\begin{align*}
\min_{G}\max_{D} V(D,G) = E_{x \sim P_{data}(z)}[\log{D(x)}] + \\
	 E_{lc \sim concat(P,c) }[\log{1-D(G(lc))}]		\tag{2} \label{eq:2}
\end{align*}

For learning the control over latent distribution additional mutual information based objective inspired by(Chen et al. 2016) is used,

\[
	L_{latentGAN} = V(D,G) - \lambda I(c;G(p,c)) \tag{3} \label{eq:3}
\]
by defining variational lower bound over second term of the equation \eqref{eq:3} $L_{latentGAN}$ can be further written as:
\[
	L_{latentGAN} = V(D,G) - \lambda L_{1}(G,Q) \tag{3} \label{eq:4}
\]
$\lambda$ term insures that GAN loss and differential entropy loss are on same scale.
    \section{Implementation Details}
We train autoencoder which consist of encoder $Enc$ which models $p(z|X)$, where $z$ is latent distribution and $X$ is the input image and decoder $Dec$ which models $p(X|z)$ using equation \eqref{eq:1}. LatentGAN generator $G$ and discriminator $D$ as shown in Figure \ref{fig:latentgan} can be implemented using linear layer with $relu$ non-linearity.The $Q$ network also has linear layer,and shares parameters with $D$ .Then $G$,$D$ and $Q$ are trained using follow process:
\begin{itemize}
\item 1) Random Gaussian noise $P$ is sampled from $N(0,I)$ along with latent code, which can be uniformly continuous $U$  or categorically discrete $Cat(K)$ where $K$ is the number of categories. 
\item 2) Latent code and Gaussian noise $P$ are passed through $G$, which generates $G(P,c)$ where $c$ can be $U$ or $Cat(K)$ code.
\item 3) $D$ gets input from $Enc(X) \sim p_{data}(z)$ and $G(P,c)$ , which outputs whether the sample belongs to $p_{data}(z)$ or not.
\item 4) $Q$ receives input from $D$, and outputs code $c$,which will be used for optimizing equation \eqref{eq:4}.
\end{itemize}
It has been observed that using layer initializations from DCGAN \cite{radford2016unsupervised} hurts LatentGAN training process,but the suggestion of using noisy labels for training $D$ helps during training.

\section{Experiments}
\subsection{MNIST Dataset}
\begin{figure}[hbt!]
	\centering
	\includegraphics[scale=0.5]{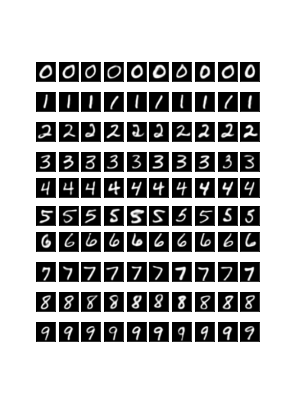}
	\caption{Varying $Cat(K)$ changes the digit category.Every row coressponds to different $K$, and every column have same $K$ but different Gaussian noise $P$. }
	\label{fig:mnist_constant_cat}
\end{figure}
\begin{figure}[hbt!]
	\centering
	\includegraphics[scale=0.5]{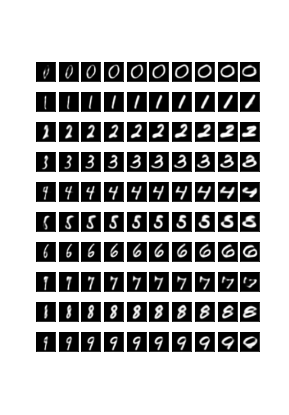}
	\caption{Varying $U_{1}(-1.5,1.5)$ changes the thickness of the digit.Every row have different $K$ and Gaussian noise $P$,and every column have different $U_{1}$ but same Gaussian noise $P$ and $K$.}
	\label{fig:mnist_uniform1_thickness}
\end{figure}
\begin{figure}[hbt!]
	\centering
	\includegraphics[scale=0.5]{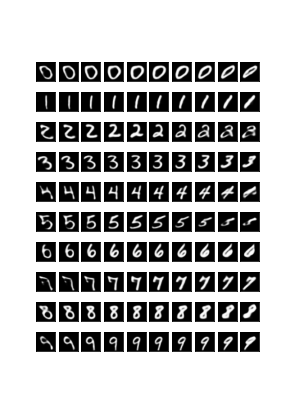}
	\caption{Varying $U_{2}(-1.5,1.5)$ rotates the digit.Every row have different $K$ and Gaussian noise $P$,and every column have different $U_{2}$ but same Gaussian noise $P$ and $K$.}
	\label{fig:mnist_uniform2_rotate}
\end{figure}
In MNIST \cite{lecun-mnisthandwrittendigit-2010} dataset, we use one categorical discrete variable  $Cat(K)$ with $K=10$ , and two  continous uniform variable $U_{1}(-1,1)$ and $U_{2}(-1,1)$ as an input latent code to $G$. After training, random samples are generated by choosing $Cat(k)$ , $P$ , $U_{1}$ and $U_{2}$ and output is shown in the Figure \ref{fig:mnist_constant_cat}.This shows that LatentGAN generator $G$ is able to approximate latent distribution of autoencoder directly.
\begin{table}[hbt!]
\centering
\begin{tabular}{|c|c|c|}
\hline 
Method & K & Test error (↓)\\
\hline
InfoGAN & 10 & 5 ± 0.01  \\
AAE & 32 & 4.10 ± 1.13 \\
\textbf{ours} & 10 & \textbf{2.38 ± 0.38} \\
\hline
\end{tabular}
\caption{MNIST Unsupervised Classification test error.}
\label{tab:mnisttesterror}
\end{table}
We have also shown learned disentangled rotation control as shown in the Figure \ref{fig:mnist_uniform2_rotate}, and disentangled thickness control in the Figure \ref{fig:mnist_uniform1_thickness}.This shows that we can directly learn control on latent distribution of autoencoder.Also we can use discriminator $D$ for unsupervised classification, and Table \ref{tab:mnisttesterror} shows that our method has better test error than InfoGAN and AAE, and (↓) denotes lower the score better the performance.
\begin{table}[hbt!]
	\centering
	\scalebox{0.6}{
	\begin{tabular}{|c|c|c|c|}
		\hline 
			Encoder $Enc$ & Decoder $Dec$ & Discriminator $D$ / $Q$ & Generator $G$\\
		\hline
			Input(1X32x32) & Input $\in R^{64}$ & Input $\in R^{64}$ & Input $\in R^{76}$ \\
		\hline
			c4-o16-s2-r & r-u4-c3-o128-p1-bn128-r & fc-o1000-r & fc-o1000-r \\
		\hline
			c4-o32-s2-r & u2-c3-o64-p1-bn64-r & fc-o1000-r & fc-o1000-r \\
		\hline
			c4-o64-s2-r & u2-c3-o32-p1-bn32-r & fc-o512-r & fc-o1000-r \\
		\hline
			c4-o128-s2-r & u2-c3-o3-p1-tanh & fc-o1-sig,fc-o2,fc-o10 & fc-o64 \\
		\hline
			c4-o128-s2-r & - & - & - \\
		\hline
			fc-o64 & - & - & - \\
		\hline
			
	\end{tabular}}
	\caption{MNIST Network Architecture}
	\label{tab:mnistnetworkarch}
\end{table}
For training LatentGAN Autoencoder on MNIST dataset, we have used architecture as mentioned in the Table \ref{tab:mnistnetworkarch} with \textit{batchsize} of \textit{128}, \textit{lr} of \textit{0.0002} and \textit{Adam} optimizer with \textit{beta1} of \textit{0.5} and \textit{beta2} of \textit{0.9} for both autoencoder and PriorGAN. In the Table \ref{tab:mnistnetworkarch} following are the conventions used, \textit{c} is \textit{convolution}, \textit{o} is \textit{output channels}, \textit{s} is \textit{stride} , \textit{u} is \textit{bilinear upsampling}, \textit{p} is \textit{padding} , \textit{r} is \textit{relu}, \textit{sig} is \textit{sigmoid} and \textit{bn} is \textit{batchnorm}. In column 3 of the table final output have 3 sections, 1st section belongs to discriminator $D$ and other sections belong to $Q$ network. $\lambda_{cont}$ and $\lambda_{disc}$ is set to 1 and 0.1 respectively.

\subsection{3D Chair Dataset}
\begin{figure}[htb!]
	\centering
	\includegraphics[scale=0.3]{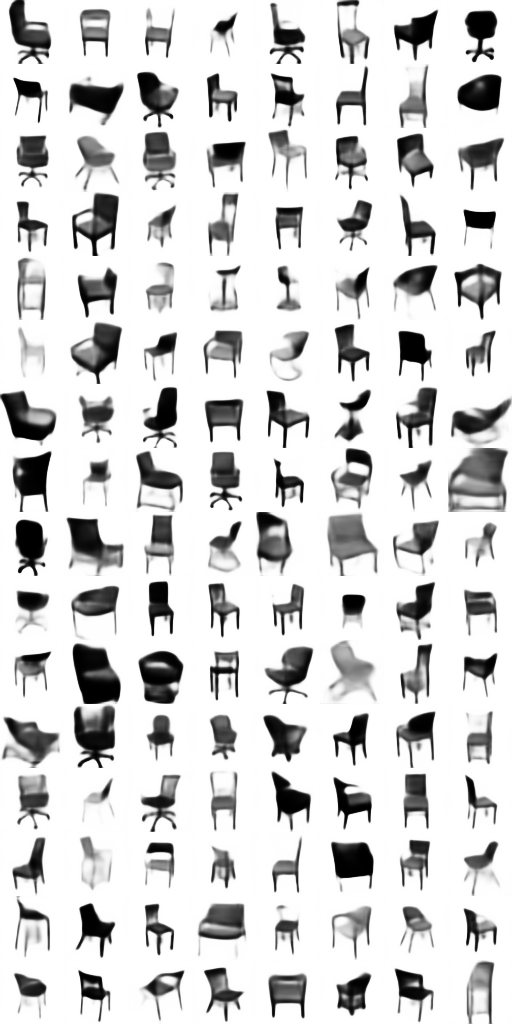}
	\caption{3D Chair Generated Samples}
	\label{fig:3dchairsamples}
\end{figure}
\begin{figure}[hbt!]
	\centering
	\includegraphics[scale=0.4]{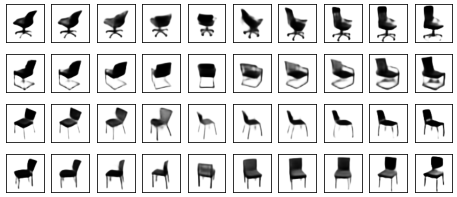}
	\caption{ Varying $U_{1}(-1,1)$ will rotate the chair. }
	\label{fig:3dchairrotate}
\end{figure}
In 3D Chair \cite{Aubry14} dataset, we use 3 discrete categorical variable $Cat(K)$ where $K=20$ and 1 continuous uniform variable $U_1(-1,1)$ as an input to $G$. We are able to generate meaningful sample as show in the Figure \ref{fig:3dchairsamples}, that means $G$ is able to approximate autoencoder latent distribution.
We are also able to learn disentangled rotational control on the chair dataset as shown in the Figure \ref{fig:3dchairrotate}. Hyperparameter setting for 3D Chair dataset is same as MNIST Dataset except $\lambda_{cont}$ and $\lambda_{disc}$ is set to 1 and 10 respectively.  

\begin{table}[hbt!]
	\centering
	\scalebox{0.6}{
	\begin{tabular}{|c|c|c|c|}
		\hline 
			Encoder $Enc$ & Decoder $Dec$ & Discriminator $D$ / $Q$ & Generator $G$\\
		\hline
			Input(1X64x64) & Input $\in R^{128}$ & Input $\in R^{128}$ & Input $\in R^{190}$ \\
		\hline
			c4-o64-s2-r & r-u4-c3-o512-p1-bn128-r & fc-o3000-r & fc-o3000-r \\
		\hline
			c4-o128-s2-r & u2-c3-o256-p1-bn64-r & fc-o3000-r & fc-o3000-r \\
		\hline
			c4-o256-s2-r & u2-c3-o128-p1-bn32-r & fc-o3000-r  & fc-o3000-r \\
		\hline
			c4-o512-s2-r & u2-c3-o64-p1-bn32-r & fc-o512-r  & fc-o3000-r \\
		\hline
			c4-o1024-s2-r & u2-c3-o1-p1-tanh& fc-o1-sig,(fc-o2),3x(fc-o20) & fc-o128\\
		\hline
			c4-o128-s2-r & - & - & \\
		\hline
			fc-o128 & - & - & - \\
		\hline
			
	\end{tabular}}
	\caption{3D Chair Network Architecture}
	\label{tab:3dchairarch}
\end{table}
For training LatentGAN Autoencoder on 3D Chair dataset, we have used architecture as mentioned in the Table \ref{tab:3dchairarch}.

\subsection{CelebA Dataset}
\begin{figure}[hbt!]
	\centering
	\includegraphics[scale=0.5]{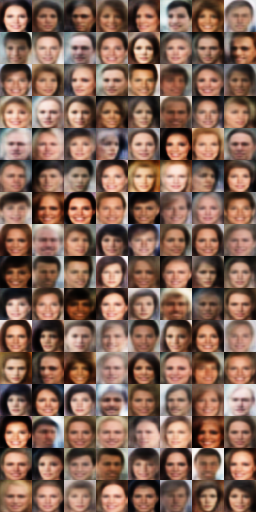}
	\caption{CelebA Generated samples}
	\label{fig:celebageneratedsamples}
\end{figure}
In CelebA \cite{liu2015faceattributes} dataset, we use 10 discrete categorical variable $Cat(K)$ only where $K=10$ as an input to $G$.After training Generator $G$  and $Dec$ are able to generate meaningful samples shown in the Figure \ref{fig:celebageneratedsamples}.
\begin{figure}[hbt!]
	\centering
	\includegraphics[scale=0.3]{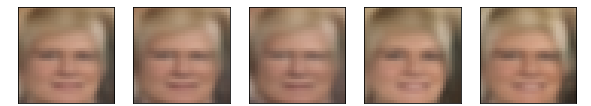}
	\caption{Varying $Cat(K)$ changes the smile.}
	\label{fig:celebasmile}
\end{figure}
Also we are able to learn smile feature control on the CelebA dataset as shown in the Figure \ref{fig:celebasmile}.

\begin{table}[hbt!]
	\centering
	\scalebox{0.6}{
	\begin{tabular}{|c|c|c|c|}
		\hline 
			Encoder $Enc$ & Decoder $Dec$ & Discriminator $D$ / $Q$ & Generator $G$\\
		\hline
			Input(3X32x32) & Input $\in R^{128}$ & Input $\in R^{128}$ & Input $\in R^{228}$ \\
		\hline
			c4-o64-s2-r & r-u4-c3-o512-p1-bn128-r & fc-o3000-r & fc-o3000-r \\
		\hline
			c4-o128-s2-r & u2-c3-o256-p1-bn64-r & fc-o3000-r & fc-o3000-r \\
		\hline
			c4-o256-s2-r & u2-c3-o128-p1-bn32-r & fc-o3000-r  & fc-o3000-r \\
		\hline
			c4-o512-s2-r & u2-c3-o3-p1-tanh & fc-o512-r  & fc-o3000-r \\
		\hline
			c4-o128-s2-r & - & fc-o1-sig,10x(fc-o10) & fc-o128\\
		\hline
			fc-o128 & - & - & - \\
		\hline
			
	\end{tabular}}
	\caption{CelebA Network Architecture}
	\label{tab:celebaarch}
\end{table}
For training LatentGAN Autoencoder on CelebA dataset, we have used architecture as mentioned in the Table \ref{tab:celebaarch}.Hyperparameter setting for CelebA dataset is same as MNIST Dataset except $\lambda_{disc}$ is set to 1.

\section{Conclusion}
We proposed LatentGAN autoencoder which can learn to control directly over autoencoder latent distribution and in doing so it is able to generate meaningful samples. Experimentally, we are able to verify that LatentGAN autoencoder can be used to learn meaningful disentangled representation over latent distribution and in unsupervised MNIST classification task performs better then InfoGAN and AAE. This further suggests that rather than making generator learn image distribution which may be challenging, we can approximate latent distribution which is less challenging and easy for generator to learn.

\section{Acknowledgement}
We thank Wobot Intelligence Inc. for providing NVIDIA GPU hardware resource for conducting this research which significantly boosted our experimentation cycle. 

\bibliography{aaai22}
\end{document}